\documentclass[sigconf]{acmart}

\AtBeginDocument{%
  \providecommand\BibTeX{{%
    \normalfont B\kern-0.5em{\scshape i\kern-0.25em b}\kern-0.8em\TeX}}}

\copyrightyear{2020} 
\acmYear{2020} 
\setcopyright{acmcopyright}\acmConference[CIKM '20]{Proceedings of the 29th ACM International Conference on Information and Knowledge Management}{October 19--23, 2020}{Virtual Event, Ireland}
\acmBooktitle{Proceedings of the 29th ACM International Conference on Information and Knowledge Management (CIKM '20), October 19--23, 2020, Virtual Event, Ireland}
\acmPrice{15.00}
\acmDOI{10.1145/3340531.3417455}
\acmISBN{978-1-4503-6859-9/20/10}

\usepackage{times}
\usepackage{latexsym}

\usepackage{hyperref}
\usepackage{url}
\usepackage{graphicx}
\usepackage{caption}
\usepackage{amsfonts}
\usepackage{amsmath}
\usepackage{multirow}
\usepackage[linesnumbered,ruled]{algorithm2e}
\DontPrintSemicolon
\usepackage{pgfplots}
\usepackage{pgfplotstable} 
\pgfplotsset{compat=newest}
\usepackage{enumitem}
\usepackage{wrapfig}

\begin{document}

\title{A Capsule Network-based Model for Learning Node Embeddings}

\author{Dai Quoc Nguyen}
\affiliation{\institution{Monash University, Australia}}
\email{dai.nguyen@monash.edu}

\author{Tu Dinh Nguyen}
\affiliation{\institution{nguyendinhtu@gmail.com}}

\author{Dat Quoc Nguyen}
\affiliation{\institution{VinAI Research, Vietnam}}
\email{v.datnq9@vinai.io}

\author{Dinh Phung}
\affiliation{\institution{Monash University, Australia}}
\email{dinh.phung@monash.edu}


\begin{abstract}
In this paper, we focus on learning low-dimensional embeddings for nodes in graph-structured data. To achieve this, we propose Caps2NE -- a new unsupervised embedding model leveraging a network of two capsule layers. Caps2NE induces a routing process to aggregate feature vectors of context neighbors of a given target node at the first capsule layer, then feed these features into the second capsule layer to infer a plausible embedding for the target node. Experimental results show that our proposed Caps2NE obtains state-of-the-art performances on benchmark datasets for the node classification task. Our code is available at: \url{https://github.com/daiquocnguyen/Caps2NE}.

\end{abstract}

\maketitle

\section{Introduction}
Numerous real-world and scientific data are represented in forms of graphs, e.g. data  from knowledge graphs, recommender systems,  social and citation networks as well as telecommunication and biological networks  \citep{battaglia2018relational,Chen180802590}. 
Recent years have witnessed many successful downstream applications of utilizing  the graph-structured data such as for improving information extraction and text classification systems \citep{kipf2017semi},  traffic  learning  and forecasting  \citep{Cui1802.07007} and for advertising and recommending relevant items to users \citep{Ying:2018:GCN,Wang:2018:BCE}. This is largely boosted by a  surge  of methodologies  that learn embedding representations to encode graph structures \citep{cai2018comprehensive}. 

One of the most important tasks in learning graph representations is to learn low-dimensional embeddings for nodes in the graph-structured data \citep{zhang2020network}. 
These embedding vectors can then be used in a downstream task such as node classification, i.e., using the learned node embeddings as feature inputs to train a classifier to predict node labels \citep{hamilton2017inductive}. 

A simple and effective approach is to treat each node as a word token and each graph as a text collection; hence we can apply a word embedding model such as Word2Vec \citep{MikolovSCCD13nips} to learn node embeddings such as DeepWalk \citep{Perozzi:2014} and Node2Vec \citep{Grover:2016}.
Recent work has developed deep neural networks (DNN) for the node classification task, e.g., GCN \citep{kipf2017semi}, GraphSAGE \citep{hamilton2017inductive} and GAT \citep{velickovic2018graph}.
We see that the DNN-based approaches are showing state-of-the-art performances, but not well-efficient to exploit the structural dependencies among nodes.

In this paper, inspired by the advanced capsule networks \citep{sabour2017dynamic}, we present Caps2NE -- a new unsupervised embedding model that adapts capsule network to learn node embeddings. Caps2NE aims to capture $h$-hops context neighbors to predict a target node. In particular, Caps2NE consists of two capsule layers with connections from the first to the second layer, but no connections within layers. 
The first layer constructs capsules to encapsulate context neighbors. 
Then a routing process is used to aggregate the feature information from capsules in the first layer to only one capsule in the second layer. 
After that, the second layer produces a continuous vector which is used to infer an embedding for the target node. 
Note that encapsulating the context neighbors into the corresponding capsules aims to preserve node properties more efficiently. 
And the routing process aims to generate high-level features to infer plausible node embeddings effectively. 

Our main contributions are as follows:

\begin{itemize}

\item We investigate the advanced use of capsule networks for the graph-structured data and propose a new embedding model Caps2NE to learn node embeddings.

\item We evaluate the performance of the proposed Caps2NE on benchmark datasets for the node classification task.

\item The experimental results show that that our Caps2NE produces state-of-the-art accuracy results on these datasets.

\end{itemize}

\begin{figure*}
\centering
\includegraphics[width=0.75\textwidth]{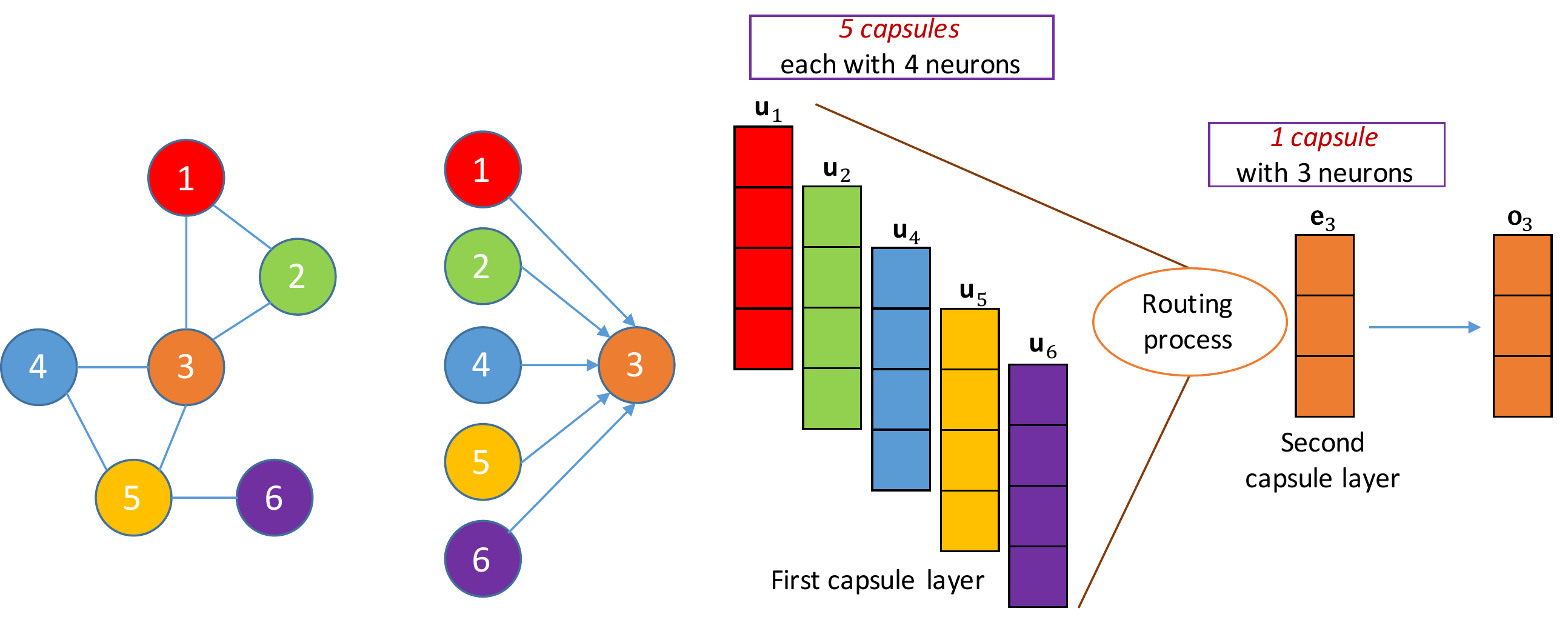}
\caption{Processes in our Caps2NE with $q = 6, d = 4, k =3$ for an illustration purpose. {Note that in this illustration, we use numbered subscripts to denote nodes themselves, not indexes of nodes or capsules.} The indexes of capsules are fixed from 1 to $(q-1)$, not depending on the indexes of the context neighbors. With $\mathsf{v}$ be the target node 3, we have $\mathsf{C}_\mathsf{v} =$ $\{\mathsf{v}_1 =$  1, $\mathsf{v}_2 =$  2, $\mathsf{v}_3 =$  4, $\mathsf{v}_4 =$  5, $\mathsf{v}_5 =$ 6$\}$.}
\label{fig:Caps2NE}
\end{figure*}

\section{The proposed Caps2NE}
\label{sec:ourmodel}

This section presents our Caps2NE model. In particular, 
we detail how to sample data from an input graph, then how to construct Caps2NE to learn node embeddings.


\textbf{Definition 1.} A network graph $\mathcal{G}$ is defined as $\mathcal{G} = (\mathcal{V}, \mathcal{E})$, in which $\mathcal{V}$ is a set of nodes, $\mathcal{E} \subseteq \{(\mathsf{v},\mathsf{v'}) | \mathsf{v}, \mathsf{v'} \in \mathcal{V}\}$ is a set of edges, and each node $\mathsf{v} \in \mathcal{V}$ is associated with a feature vector $\boldsymbol{x}_\mathsf{v} \in \mathbb{R}^{d}$.
We aim to learn a node embedding $\boldsymbol{\mathsf{o}}_\mathsf{v}$ for each node $\mathsf{v}$.



\textbf{Sampling input pairs.} We follow \citet{Perozzi:2014} to uniformly sample a number $T$ of random walks of length $q$ for every node in $\mathcal{V}$. 
From each random walk, we randomly sample a target node $\mathsf{v}$, treat $(q-1)$ remaining nodes as the context neighbors of node $\mathsf{v}$, and construct an input pair of ($\mathsf{C}_\mathsf{v}$, $\mathsf{v}$), where we denote $\mathsf{C}_\mathsf{v}$ be the list of context neighbors $\mathsf{v}_i$ of the target node $\mathsf{v}$ (here, $i \in \{1, 2, ..., q-1\}$ and $|\mathsf{C}_\mathsf{v}| = q-1$). 

Figure \ref{fig:Caps2NE} shows an example of a graph consisting of 6 nodes.
If we sample a random walk of length $q = 6$ for node $1$ such as $\{1, 2, 3, 4, 5, 6\}$ and select node $3$ as the target node $\mathsf{v}$, then the remaining nodes $\{1, 2, 4, 5, 6\}$ are  treated as the context neighbors of node $3$, i.e., $\mathsf{C}_\mathsf{v} =$ $\{\mathsf{v}_1 =$  1, $\mathsf{v}_2 =$  2, $\mathsf{v}_3 =$  4, $\mathsf{v}_4 =$  5, $\mathsf{v}_5 =$ 6$\}$.

\textbf{Definition 2.} A \textit{capsule} is a group of neurons. A capsule \textit{layer} is a group of capsules without connections among capsules in the same layer \citep{sabour2017dynamic}. Two continuous capsule layer is connected using a \textit{routing} process.

\textbf{Constructing Caps2NE.} We build our Caps2NE with two capsule layers. In the first layer, we construct $(q-1)$ capsules, where the feature vector of each context neighbor $\mathsf{v}_i$ is encapsulated by the $i$-th corresponding capsule (with $i \in \{1, 2, ..., q-1\}$). 
In the second layer, we construct one capsule to produce a vector representation which is then used to infer an embedding for the target node $\mathsf{v}$.

The first capsule layer consists of $(q-1)$ capsules, in which the $i$-th capsule use a non-linear squashing function to transform the feature vector $\boldsymbol{x}_{\mathsf{v}_i}$ of the context neighbor $\mathsf{v}_i$ into $\boldsymbol{\mathsf{u}}_{\mathsf{v}_i}^{(i)}$ as: 
\begin{equation}
    \boldsymbol{\mathsf{u}}_{\mathsf{v}_i}^{(i)} = \mathsf{squash}\left(\boldsymbol{x}_{\mathsf{v}_i}\right) = \frac{\|\boldsymbol{x}_{\mathsf{v}_i}\|^2}{1 + \|\boldsymbol{x}_{\mathsf{v}_i}\|^2}\frac{\boldsymbol{x}_{\mathsf{v}_i}}{\|\boldsymbol{x}_{\mathsf{v}_i}\|}
\label{equa:squash}
\end{equation}
\noindent The squashing function ensures that the orientation of each feature vector is unchanged while its length is scaled down to below 1.

Vectors $\boldsymbol{\mathsf{u}}_{\mathsf{v}_i}^{(i)}$ are then linearly transformed using weight matrices $\textbf{W}_{i} \in \mathbb{R}^{k\times d}$ to produce vectors $\hat{\boldsymbol{\mathsf{u}}}_{\mathsf{v}_i}^{(i)} \in \mathbb{R}^{k}$. 
These vectors $\hat{\boldsymbol{\mathsf{u}}}_{\mathsf{v}_i}^{(i)}$  are weighted to sum up to return a vector  $\boldsymbol{\mathsf{s}}_{\mathsf{v}} \in \mathbb{R}^{k}$ for the capsule in the second layer (recall that the second layer consists of only one  capsule).
This capsule then performs the non-linear squashing function to produce a vector $\boldsymbol{\mathsf{e}}_{\mathsf{v}} \in \mathbb{R}^{k}$. 
Formally, we have:
\begin{eqnarray}
\boldsymbol{\mathsf{e}}_{\mathsf{v}} = \mathsf{squash}\left(\boldsymbol{\mathsf{s}}_{\mathsf{v}}\right) \ \  ;   \ \ \boldsymbol{\mathsf{s}}_{\mathsf{v}} = \sum_{i} c_{i}\hat{\boldsymbol{\mathsf{u}}}_{\mathsf{v}_i}^{(i)} \ \ ;  \ \  \hat{\boldsymbol{\mathsf{u}}}_{\mathsf{v}_i}^{(i)} = \textbf{W}_{i} \boldsymbol{\mathsf{u}}_{\mathsf{v}_i}^{(i)} 
\end{eqnarray}
where $c_{i}$ are coupling coefficients determined by the routing process as presented in Algorithm \ref{alg:routing}. Here, 
$c_{i}$ aims to weight $\boldsymbol{\mathsf{u}}_{\mathsf{v}_i}^{(i)}$ of the $i$-th capsule in the first layer. 

As we use one capsule in the second layer, we make two differences in our routing process  in Algorithm \ref{alg:routing}: (i) we apply  $\mathsf{softmax}$ in a direction from all capsules in the previous layer to each of capsules in the next layer, 
(ii) thus, we propose a new update rule ($b_{i} \leftarrow \hat{\boldsymbol{\mathsf{u}}}_{\mathsf{v}_i}^{(i)}\cdot\boldsymbol{\mathsf{e}}_{\mathsf{v}}$) instead of employing ($b_{i} \leftarrow b_{i} + \hat{\boldsymbol{\mathsf{u}}}_{\mathsf{v}_i}^{(i)}\cdot\boldsymbol{\mathsf{e}}_{\mathsf{v}}$) originally used by  \citet{sabour2017dynamic}.

\begin{algorithm}
\caption{The Caps2NE routing process.}
\label{alg:routing}
\DontPrintSemicolon
\SetAlgoVlined
\For{i = 1, 2, ..., q-1}{
    $b_{i} \leftarrow$ 0 
}
\For{$\mathsf{iteration}$ = 1, 2, ..., m}{
    $\boldsymbol{\mathsf{c}} \leftarrow \mathsf{softmax}\left(\boldsymbol{\mathsf{b}}\right)$  
    
    $\boldsymbol{\mathsf{s}}_{\mathsf{v}} \leftarrow \sum_{i} c_{i}\hat{\boldsymbol{\mathsf{u}}}_{\mathsf{v}_i}^{(i)}$
    
    $\boldsymbol{\mathsf{e}}_{\mathsf{v}} \leftarrow \mathsf{squash}\left(\boldsymbol{\mathsf{s}}_{\mathsf{v}}\right)$
    
    \For{i = 1, 2, ..., q-1}{
        $b_{i} \leftarrow \hat{\boldsymbol{\mathsf{u}}}_{\mathsf{v}_i}^{(i)}\cdot\boldsymbol{\mathsf{e}}_{\mathsf{v}}$ 
}
}
\end{algorithm}

\textbf{Learning model parameters.} The vector representation  $\boldsymbol{\mathsf{e}}_{\mathsf{v}}$ is then used to infer the final embedding $\boldsymbol{\mathsf{o}}_{\mathsf{v}} \in \mathbb{R}^{k}$ of the target node $\mathsf{v}$, as shown in Equation \ref{equa:Caps2NE}.
We learn all model parameters (including the node embeddings $\boldsymbol{\mathsf{o}}_{\mathsf{v}}$) by minimizing the sampled softmax loss function \citep{Jean2015} applied to the target node $\mathsf{v}$ as:
\begin{equation}
\mathcal{L}_{\mathsf{Caps2NE}}\left(\mathsf{v}\right) = -\log \frac{\exp(\boldsymbol{\mathsf{o}}_\mathsf{v}^\mathsf{T} \boldsymbol{\mathsf{e}}_{\mathsf{v}})}{\sum_{\mathsf{v'} \in \mathcal{V'}} \exp(\boldsymbol{\mathsf{o}}_\mathsf{v'}^\mathsf{T} \boldsymbol{\mathsf{e}}_{\mathsf{v}})} 
\label{equa:Caps2NE}
\end{equation}
\noindent where $\mathcal{V'}$ is a subset sampled from $\mathcal{V}$.

\begin{algorithm}
\caption{The Caps2NE learning process.}
\label{alg:Caps2NE}
{
\DontPrintSemicolon
\SetAlgoVlined
\textbf{Input}: A network graph $\mathcal{G} = (\mathcal{V}, \mathcal{E})$

\For{$\mathsf{v} \in \mathcal{V}$}{
	\textsc{Sample} $T$ random walks of length $q$ starting at $\mathsf{v}$
}

\For{each $\mathsf{random\ walk}$}{
\textsc{Sample} a node $\mathsf{v}$ as a target node

$\mathsf{C}_\mathsf{v} \leftarrow$ Remaining nodes

\For{i = 1, 2, ..., q-1}{
    $\boldsymbol{\mathsf{u}}_{\mathsf{v}_i}^{(i)} \leftarrow \mathsf{squash}\left(\boldsymbol{x}_{\mathsf{v}_i}\right) \quad \forall \mathsf{v}_i \in \mathsf{C}_\mathsf{v}$
}

$\boldsymbol{\mathsf{e}}_{\mathsf{v}} \leftarrow \textsc{Routing}\left(\left\{\boldsymbol{\mathsf{u}}_{\mathsf{v}_i}^{(i)}\right\}_{i=1}^{q-1}\right)$

$\boldsymbol{\mathsf{o}}_{\mathsf{v}} \leftarrow \boldsymbol{\mathsf{e}}_{\mathsf{v}}$
}
}
\end{algorithm}

We briefly  represent the general learning process of our proposed Caps2NE model in Algorithm \ref{alg:Caps2NE} whose main steps \textbf{3}, \textbf{7--9} and \textbf{10} are previously detailed in parts ``\textit{Sampling input pairs}'', ``\textit{Constructing Caps2NE}'' and ``\textit{Learning model parameters}'', respectively. 

We illustrate our model in Figure \ref{fig:Caps2NE} where the length $q$ of random walks, the dimension size $d$ of the feature vectors and the  dimension size $k$ of output node embeddings are equal to 6, 4 and 3, respectively. 
Thus, the first capsule layer has 5 capsules, each with 4 neurons, and the second capsule layer has 1 capsule with 3 neurons. 
For the target node $3$ in the illustration, the vector output of the capsule in the second layer is used to infer the embedding of node $3$.
Our Caps2NE aims to aggregate feature information from the context neighbors (i.e., $k$-hops neighbors) to infer the target node 3; hence this helps our proposed model to infer the structural dependencies among nodes to produce the plausible node embeddings effectively.

\begin{algorithm}
\caption{The inference process for new nodes.}
\label{alg:InferInd}
{
\DontPrintSemicolon
\SetAlgoVlined
\textbf{Input}: A graph $\mathcal{G} = (\mathcal{V}, \mathcal{E})$, a trained model Caps2NE$_{trained}$, a set $\mathcal{V}_{test}$ of new nodes.

\For{$v \in \mathcal{V}_{test}$}{
	\textsc{Sample} $Z$ pairs \{\textit{p}$_j$\}$_{j=1}^Z$ of ($\mathsf{C}_v$, $v$)
    
\For{$j \in \{1,\ 2,\ ...,\ Z\}$}{
    $\boldsymbol{\mathsf{e}}_{(v,j)} \leftarrow$ Caps2NE$_{trained}\left(\textit{p}_j\right)$
}

$\boldsymbol{\mathsf{o}}_v \leftarrow \textsc{Average}\left(\{\boldsymbol{\mathsf{e}}_{(v,j)}\}_{j=1}^Z\right)$

}
}
\end{algorithm}

\textbf{Inferring embeddings for new nodes in the inductive setting.}
Algorithm \ref{alg:InferInd} shows how we infer an embedding for a new node $v$ adding to an existing graph. 
After training our model, we generate random walks of length $q$ to extract $Z$ pairs of ($\mathsf{C}_v$, $v$).
We use each of these pairs as an input for our trained model and then collect the output vector $\boldsymbol{\mathsf{e}}$ from the second capsule layer.
Thus, we obtain $Z$ vectors associated with node $\mathsf{v}$ and then average them into an embedding representation of $v$.

\section{Experimental results on PPI, POS, and \textsc{BlogCatalog}}

\subsection{Datasets and data splits}

PPI \citep{BioGRID:2008} is a subgraph of the Protein-Protein Interaction network for Homo Sapiens, and its node labels represent biological states. POS \citep{mahoney2011large} is a co-occurrence network of words from the Wikipedia dump, and its node labels represent the part-of-speech tags. \textsc{BlogCatalog} \citep{ZafaraniLiu:2009} is a social network of relationships of the bloggers listed on the BlogCatalog website, and its node labels represent bloggers' interests.  {PPI, POS} and \textsc{BlogCatalog} are given without node features, in which each node is assigned with one or more class labels. These datasets are used for the multi-label node classification task.
Table \ref{tab:graphdatasets_pos} presents the statistics of these benchmark datasets.

\begin{table}[!ht]
\centering
\caption{Statistics of the experimental datasets.}
\def\arraystretch{1.1}
\begin{tabular}{l|llc}
\hline
\bf Dataset &  \bf{$|\mathcal{V}|$} & \bf{$|\mathcal{E}|$} & {\#Classes}\\
\hline
PPI & 3,890 & 76,584 & 50 \\
POS & 4,777 & 184,812 & 40 \\
\textsc{BlogCatalog} & 10,312 & 333,983 & 39\\
\hline
\end{tabular}
\label{tab:graphdatasets_pos}
\end{table}

A certain fraction $\gamma$ of nodes is provided to train a classifier which is then used to predict the labels of the remaining nodes.

\subsection{Training protocol}

We only use the transductive setting for these three datasets. We uniformly sample 64 random walks ($T = 64$) of length 10 ($q= 10$) for each node in the graph. 
In each random walk, we rotationally select each node in the walk as a target node and 9 remaining nodes as its context nodes.
We also run up to 50 training epochs and use the batch size to 128, the embedding size $k= 128$ and  $|\mathcal{V'}| = 256$ in Equation \ref{equa:Caps2NE}.
We vary the Adam initial learning rate $lr \in \{1e^{-5}, 5e^{-5}, 1e^{-4}\}$. 
Nodes are given without pre-computed features, hence we set the  size $d$ of feature vectors $\boldsymbol{x}_{\mathsf{v}_i}$ to 128 ($d = 128$), and these vectors are randomly  initialized uniformly, and updated during training.

\subsection{Evaluation protocol}

We follow the same experimental setup used for the multi-label node classification task from \citet{Perozzi:2014} and \citet{duran2017learning} where we uniformly sample a fraction $\gamma$ of nodes at random as training set for learning a one-vs-rest logistic regression classifier. 
The learned node embeddings after each Caps2NE training epoch are used as input feature vectors  for this logistic regression classifier. We use default parameters for learning this classifier from \citet{Perozzi:2014}. 
The classifier is then used to categorize the remaining nodes.
We monitor the Micro-F1 and Macro-F1 scores of the classifier after  each Caps2NE training epoch, for which the best model is chosen by using 10-fold cross-validation for each fraction value. 
We repeat this manner 10 times for each fraction value, and then compute the averaged Micro-F1 and Macro-F1 scores. We show final scores w.r.t. each value  $\gamma \in \{10\%, 50\%, 90\%\}$.
The baseline results are taken from \citet{duran2017learning}.

\subsection{Overall results}

\begin{table*}
\centering
\caption{Multi-label classification results on PPI, POS and \textsc{BlogCatalog}.}
\def\arraystretch{1.2}
\begin{tabular}{l|ccc|ccc|ccc}
\hline
\bf Method & \multicolumn{3}{c|}{\bf POS} & \multicolumn{3}{c|}{\bf PPI} & \multicolumn{3}{c}{\bf \textsc{BlogCatalog}}\\
\cline{2-10}
(Micro-F1) & $\gamma=10\%$ & $\gamma=50\%$ & $\gamma=90\%$ & $\gamma=10\%$ & $\gamma=50\%$ & $\gamma=90\%$ & $\gamma=10\%$ & $\gamma=50\%$ & $\gamma=90\%$\\
\hline
DeepWalk &  45.02 &  49.10 &  49.33  & 17.14   & \textbf{23.52}  & 25.02  & 34.48   & 38.11   & 38.34 \\
LINE  & 45.22  & \textbf{51.64}  & \underline{52.28}  & 16.55   & 23.01   & \textbf{25.28} & 34.83   & 38.99   & 38.77 \\
Node2Vec  & 44.66   & 48.73   & 49.73   & 17.00  & \underline{23.31} & 24.75  & \textbf{35.54}   & \underline{39.31}  & 40.03 \\
EP-B  & \textbf{46.97}  &  49.52   & 50.05   & \underline{17.82} & 23.30  &  24.74  & \underline{35.05}   & \textbf{39.44}  & \underline{40.41}\\
\hline
Our Caps2NE  & \underline{46.01} & \underline{50.93} & \textbf{53.92} & \textbf{18.52} & 23.15  & \underline{25.08} & 34.31  & 38.35  & \textbf{40.79}\\
\hline
\hline
{\bf Method}& \multicolumn{3}{c|}{\bf POS} & \multicolumn{3}{c|}{\bf PPI}  &\multicolumn{3}{c}{\bf \textsc{BlogCatalog}}\\
\cline{2-10}
(Macro-F1) & $\gamma=10\%$ & $\gamma=50\%$ & $\gamma=90\%$ & $\gamma=10\%$ & $\gamma=50\%$ & $\gamma=90\%$ & $\gamma=10\%$ & $\gamma=50\%$ & $\gamma=90\%$\\
\hline
DeepWalk & 8.20   & 10.84   & 12.23  & 13.01  &  18.73  & 20.01  & 18.16  &  22.65   & 22.86 \\
LINE & 8.49   & \underline{12.43}  & \underline{12.40}  & 12,79   & 18.06  &  \textbf{20.59} & 18.13   & 22.56  &  23.00 \\
Node2Vec & 8.32  &  11.07   & 12.11  &  13.32   & 18.57   & 19.66  & \textbf{19.08}  & 23.97   & 24.82 \\
EP-B & \underline{8.85} & 10.45   & 12.17  &  \underline{13.80} & \underline{18.96} & \underline{20.36} & \textbf{19.08}  &  \textbf{25.11}  & \underline{25.97} \\
\hline
Our Caps2NE & \textbf{9.71} & \textbf{13.16} & \textbf{14.11} & \textbf{15.20} & \textbf{19.63} & 20.27  & \underline{18.40}   & \underline{24.80} & \textbf{26.63}\\
\hline
\end{tabular}
\label{tab:posppi}
\end{table*}

We show in Table \ref{tab:posppi} the Micro-F1 and Macro-F1 scores on test sets in the transductive setting. Especially, on POS, Caps2NE produces a new state-of-the-art Macro-F1 score for each of the three fraction values $\gamma$, the highest Micro-F1 score when $\gamma = 90\%$ and the second highest Micro-F1  scores when  $\gamma \in \{10\%, 50\%\}$. Caps2NE   obtains new highest F1 scores on PPI and \textsc{BlogCatalog} when $\gamma = 10\%$ and $\gamma = 90\%$, respectively. On PPI, Caps2NE also achieves the highest Macro-F1 score  when $\gamma = 50\%$ and the second highest Micro-F1 score  when $\gamma = 90\%$. On \textsc{BlogCatalog}, Caps2NE also achieves the second highest Macro-F1 scores  when   $\gamma \in \{10\%, 50\%\}$. 

In short, from Table \ref{tab:posppi}, Caps2NE obtains top performances on these  three datasets: producing the highest scores in 9 over 18 comparison groups (3 datasets $\times$ 3 values of the fraction $\gamma$ $\times$ 2 metrics), the second  highest scores in 5/18 groups and competitive scores in the remaining 4 groups.

\section{Experimental results on \textsc{Cora, Citeseer}, and \textsc{Pubmed}}

\subsection{Datasets and data splits}

\textsc{Cora, Citeseer} \citep{sen2008collective} and \textsc{Pubmed} \citep{namata:mlg12} are citation networks where each node represents a document (here, each node is associated with a class labeling the main topic of the document), and each edge represents a citation link between two documents.  
Each node is also associated with a feature vector of a bag-of-words, i.e. the feature vectors $\boldsymbol{x}_{\mathsf{v}_i}$ in the first capsule layer (Equation \ref{equa:squash}) are pre-computed based on bag-of-words features and fixed during training.
Table \ref{tab:graphdatasets} presents the statistics of these three benchmark datasets.

\begin{table}[!ht]
\centering
\caption{Statistics of the experimental datasets. $d$ is the dimension size of the feature vectors.}
\def\arraystretch{1.1}
\begin{tabular}{l|llcl}
\hline
\bf Dataset &  \bf{$|\mathcal{V}|$} & \bf{$|\mathcal{E}|$} & {\#Classes} & $d$\\
\hline
\textsc{Cora} & 2,708 & 5,429 & 7 & 1,433\\
\textsc{Citeseer} & 3,327 & 4,732 & 6 & 3,703\\
\textsc{Pubmed} & 19,717 & 44,338 & 3 & 500\\
\hline
\end{tabular}
\label{tab:graphdatasets}
\end{table}

\citet{duran2017learning} show that the experimental setup used in \citep{kipf2017semi,velickovic2018graph} is not fair to show the effectiveness of existing models when these models are evaluated using the fixed \& pre-split training, validation and test sets from the Planetoid model \citep{Yang:2016planetoid}.
Therefore, for a fair comparison, we follow the same experimental setup used in \citep{duran2017learning,Nguyen2019SANNE}. 
In particular, for each dataset, we uniformly sample 20 random nodes for each class as training data, 1000 different random nodes as a validation set and 1000 different random nodes as a test set.
We then repeat this manner 10 times to produce 10 data splits of training-validation-test sets.

\subsection{Training protocol}
\label{subsec:train}

\textbf{Transductive setting.} We set the embedding size $k$ to 128 ($k = 128$) and the number of samples in the sampled softmax loss function to 256 ($|\mathcal{V'}| = 256$ in Equation \ref{equa:Caps2NE}).
We also set the batch size to 64 for both \textsc{Cora} and \textsc{Citeseer} and to 128 for \textsc{Pubmed}.
We use a fixed walk length $q$ = 10 for uniformly sampling $T$ random walks starting from each node.
We may get slightly better results when we rotationally selecting each node in the random walk as a target node. 
But we aim to save training time due to the limitation of computation resources, thus we only select target nodes at indexes of $\{3, 4, 5, 6\}$.
We optimize the loss function using the Adam optimizer \citep{kingma2014adam} and select the initial learning rate $lr \in \{1e^{-5}, 5e^{-5}, 1e^{-4}\}$.
We vary the number $T$ of random walks $T \in \{8, 16, 32, 64\}$  and the number $m$ of iterations in the routing process (Algorithm \ref{alg:routing}) $m \in \{1, 3, 5, 7\}$.
We run up to 50 epochs and evaluate the model for each epoch to choose the best model on the validation set.
We use the same values of hyper-parameters above for all data splits.

\textbf{Inductive setting.} We use the same inductive setting as used in  \citep{Yang:2016planetoid,duran2017learning} where \textit{we firstly remove all nodes in the test set from the original graph before training phase, thus these nodes  are unseen/new in the testing/evaluating phase.} We then apply the standard training process on the remaining of the graph. 
Here, we use the same set of hyper-parameters tuned for the transductive setting to train Caps2NE in the inductive setting.
After training, we infer the embedding for each node $\mathsf{v}$ in the test set as  in Algorithm \ref{alg:InferInd} using a fixed value $Z=10$.

\subsection{Evaluation protocol}
\label{subsec:eval}

We also follow the same setup used in \citet{duran2017learning} use to evaluate our Caps2NE. 
For each of 10 data splits, the learned node embeddings after each Caps2NE training epoch are used as input features for learning a L2-regularized logistic regression classifier \citep{Fan:2008}
on the training set.
We monitor the node classification accuracy on the validation set for every Caps2NE training epoch and then choose the model that produces the highest accuracy on the validation set to compute the accuracy on the test set. 
We finally report the average of the accuracies across 10 test sets from the 10 data splits.
We compare Caps2NE with strong baseline models  BoW (Bag-of-Words), DeepWalk, DeepWalk+BoW, EP-B \citep{duran2017learning}, Planetoid, GCN and GAT.
As reported in \citep{guo2018spine}, GraphSAGE obtained low accuracies on \textsc{Cora}, \textsc{Pubmed} and \textsc{Citeseer}, thus we do not include GraphSAGE as a strong baseline.

\subsection{Overall results}

\textbf{Transductive setting.} Table \ref{tab:expresults} reports the experimental results of our proposed Caps2NE and other baselines. BoW is evaluated by directly using the bag-of-words feature vectors for learning the classifier. 
DeepWalk+BoW concatenates the learned embedding of a node from DeepWalk with the BoW feature vector of the node.
As discussed in \citet{duran2017learning}, the experimental setup used to evaluate GCN  and GAT  is not fair for existing models when they are evaluated using the fixed \& pre-split training, validation and test sets from \citet{Yang:2016planetoid}.
Thus we report results, and also  fine-tune and re-evaluate GAT,  
using the same experimental setup used in \citet{duran2017learning}.
The results of other baselines (e.g., BoW, DeepWalk+BoW, EP-B, Planetoid and GCN) are taken from \citet{duran2017learning}.

\begin{table}[!ht]
\centering
\caption{Accuracies on the \textsc{Cora}, \textsc{Citeseer} and \textsc{Pubmed} {test} sets in the transductive setting.
``\textbf{Unsup}'' denotes unsupervised graph embedding models, where the best score is in {bold} while the second best score is in {underline}. ``\textbf{Semi}'' denotes a group of semi-supervised models using node labels from the training set together with feature vectors of nodes from the entire dataset during training.
}
\def\arraystretch{1.1}
\begin{tabular}{l|l|c|c|c}
\hline
\multicolumn{2}{c|}{\bf Model} &  \textbf{\textsc{Cora}} & \textbf{\textsc{Citeseer}} & \textbf{\textsc{Pubmed}} \\
\hline
\multirow{5}{*}{\rotatebox[origin=c]{90}{\textbf{Unsup}}} & BoW & 58.63  & 58.07  & 70.49  \\
& DeepWalk  & 71.11  & 47.60  & 73.49  \\
& DeepWalk+BoW & 76.15  & 61.87  & 77.82 \\
& EP-B  & \underline{78.05}  & \underline{71.01} & \textbf{79.56}\\
& Our \textbf{Caps2NE} & \textbf{80.53} & \textbf{71.34} & \underline{78.45}  \\
\hline
\multirow{3}{*}{\rotatebox[origin=c]{90}{\textbf{Semi}}} & GAT  & 81.72 & 70.80  & {79.56}\\
& GCN & 79.59  & 69.21  & {77.32}\\
& Planetoid   & 71.90  & 58.58  & 74.49  \\
\hline
\end{tabular}
\label{tab:expresults}
\end{table}

Caps2NE obtains the highest scores on \textsc{Cora} and \textsc{Citeseer} and the second highest score on \textsc{Pubmed} against other unsupervised baseline models.
In addition, we also compare our unsupervised Caps2NE to the semi-supervised models GCN, Planetoid and GAT, for which Caps2NE works better than GCN and Planetoid on these three datasets, and outperforms GAT on \textsc{Citeseer}.

\begin{table}[!ht]
\centering
\caption{Accuracies on the \textsc{Cora}, \textsc{Citeseer} and \textsc{Pubmed} {test} sets in the inductive setting.
``\textbf{Unsup}'' denotes unsupervised graph embedding models, where the best score is in {bold} while the second best score is in {underline}. ``\textbf{Sup}'' denotes a group of supervised models using node labels from the training set during training.
}
\def\arraystretch{1.1}
\begin{tabular}{l|l|c|c|c}
\hline
\multicolumn{2}{c|}{\bf Model} &  \textbf{\textsc{Cora}} & \textbf{\textsc{Citeseer}} & \textbf{\textsc{Pubmed}} \\
\hline
\multirow{3}{*}{\rotatebox[origin=c]{90}{\textbf{Unsup}}} 
& DeepWalk+BoW & 68.35  & 59.47  & 74.87 \\
& EP-B  & \underline{73.09} & \underline{68.61} & \textbf{79.94}\\
& Our \textbf{Caps2NE} & \textbf{76.54} & \textbf{69.84} & \underline{78.98} \\
\hline
\multirow{3}{*}{\rotatebox[origin=c]{90}{\textbf{Sup}}} & GAT & {69.37} & 59.55  & {71.29}\\
& GCN  & 67.76  & 63.40  & 73.47 \\
& Planetoid  & 64.80  & 61.97  & {75.73}\\
\hline
\end{tabular}
\label{tab:expresults_ind}
\end{table}

\textbf{Inductive setting:} 
Table \ref{tab:expresults_ind} reports the experimental results of our Caps2NE and other baselines in the inductive setting.
Note that the inductive setting is used to evaluate the models when we do not access nodes in the test set during training.
This inductive setting was missed in the original GCN and GAT papers which relied on the semi-supervised training process.
Regarding Cora and Citeseer in the inductive setting, many neighbors of test nodes also belong to the test set, thus these neighbors are unseen during training and then become new nodes in the testing/evaluating phase.
Table \ref{tab:expresults} also shows that  under the inductive setting, Caps2NE produces new state-of-the-art scores of 76.54\% and 69.84\% on \textsc{Cora} and \textsc{Citeseer} respectively, and also obtains the second highest score of 78.98\% on \textsc{Pubmed}. As previously discussed in the last paragraph in the ``The proposed Caps2NE'' section, we re-emphasize that our unsupervised Caps2NE model notably outperforms the supervised models GCN and GAT for this inductive setting.
In particular, Caps2NE achieves 4+\% absolute higher accuracies than both GCN and GAT on the three datasets,  clearly showing the effectiveness of Caps2NE to infer embeddings for unseen nodes.

\textbf{Discussion.} EP-B is the best model on \textsc{Pubmed}: (i) EP-B simultaneously learns word embeddings on texts from all nodes. Then the embeddings of words from each node are averaged into a new feature vector which is then used to reconstruct the node embedding.
(ii) On \textsc{Pubmed}, neighbors of unseen nodes in the test set are frequently present in the training set.
Therefore, these are reasons why on \textsc{Pubmed}, EP-B obtains higher performance than Caps2NE and other models (but, note that we only make use of the bag-of-words feature vectors).

\begin{figure*}[!ht]
\centering
\centering
\resizebox{16.75cm}{!}{
\begin{tikzpicture}
\begin{axis}[
    ybar,
    enlarge x limits=0.25,
    legend style={at={(0.5,1)},
                anchor=north,legend columns=3},
    ylabel={Accuracy},
    xlabel={$lr$},
    symbolic x coords={$1e^{-5}$, $5e^{-5}$, $1e^{-4}$},
    xtick=data,
    ymin=69,ymax=84,
    nodes near coords={\scriptsize\pgfmathprintnumber\pgfplotspointmeta},
    ]
\addplot+[error bars/.cd, y dir=both,y explicit] coordinates {
($1e^{-5}$,77.90) +- (0.0, 1.63)
($5e^{-5}$,80.46) +- (0.0, 1.25) 
($1e^{-4}$,80.33) +- (0.0, 1.30)
};
\addplot+[error bars/.cd, y dir=both,y explicit] coordinates {
($1e^{-5}$,71.90) +- (0.0, 1.63)
($5e^{-5}$,72.28) +- (0.0, 1.67) 
($1e^{-4}$,72.22) +- (0.0, 1.66)
};
\addplot+[error bars/.cd, y dir=both,y explicit] coordinates {
($1e^{-5}$,73.79) +- (0.0, 2.65)
($5e^{-5}$,76.38) +- (0.0, 2.23) 
($1e^{-4}$,76.53) +- (0.0, 2.34)
};
\legend{\textsc{Cora},\textsc{Citeseer},\textsc{Pubmed}}
\end{axis}
\end{tikzpicture}

\begin{tikzpicture}
\begin{axis}[
    ybar,
    enlarge x limits=0.25,
    legend style={at={(0.5,1)},
                anchor=north,legend columns=3},
    ylabel={Accuracy},
    xlabel={$T$},
    symbolic x coords={8,16, 32, 64},
    xtick=data,
    ymin=69,ymax=84,
    nodes near coords={\scriptsize\pgfmathprintnumber\pgfplotspointmeta},
    ]
\addplot+[error bars/.cd, y dir=both,y explicit] coordinates {
(8,78.66) +- (0.0, 1.34)
(16,79.73) +- (0.0, 1.15) 
(32,80.51) +- (0.0, 1.39)
(64,79.75)+- (0.0, 1.11)
};
\addplot+[error bars/.cd, y dir=both,y explicit] coordinates {
(8,71.45) +- (0.0, 1.50)
(16,71.87) +- (0.0, 1.50) 
(32,71.88) +- (0.0, 1.73)
(64,72.03)+- (0.0, 1.71)
};
\addplot+[error bars/.cd, y dir=both,y explicit] coordinates {
(8,74.74) +- (0.0, 2.46)
(16,76.24) +- (0.0, 2.04) 
(32,76.16) +- (0.0, 2.31)
(64,76.37)+- (0.0, 2.01)
};
\legend{\textsc{Cora},\textsc{Citeseer},\textsc{Pubmed}}
\end{axis}
\end{tikzpicture}

\begin{tikzpicture}
\begin{axis}[
    ybar,
    enlarge x limits=0.25,
    legend style={at={(0.5,1)},
                anchor=north,legend columns=3},
    ylabel={Accuracy},
    xlabel={$m$},
    symbolic x coords={1, 3, 5, 7},
    xtick=data,
    ymin=69,ymax=84,
    nodes near coords={\scriptsize\pgfmathprintnumber\pgfplotspointmeta},
    ]
\addplot+[error bars/.cd, y dir=both,y explicit] coordinates {
(1,80.48) +- (0.0, 1.41)
(3,79.98) +- (0.0, 1.29) 
(5,80.21) +- (0.0, 1.15)
(7,79.96) +- (0.0, 1.32)
};
\addplot+[error bars/.cd, y dir=both,y explicit] coordinates {
(1,71.99) +- (0.0, 1.46)
(3,71.94) +- (0.0, 1.61) 
(5,72.0) +- (0.0, 1.63)
(7,72.04) +- (0.0, 1.67)
};
\addplot+[error bars/.cd, y dir=both,y explicit] coordinates {
(1,76.34) +- (0.0, 2.20)
(3,76.09) +- (0.0, 1.96) 
(5,76.23) +- (0.0, 2.08)
(7,76.22) +- (0.0, 2.13)
};
\legend{\textsc{Cora},\textsc{Citeseer},\textsc{Pubmed}}
\end{axis}
\end{tikzpicture}
}
\caption{Effects of the Adam initial  learning rate $lr$ (left figure), the number $T$ of random walks sampled for each node (central figure), and the number $m$ of iterations in the routing process (right figure) on the validation sets in the transductive setting.}
\label{fig:EffectsTransductive}
\end{figure*}
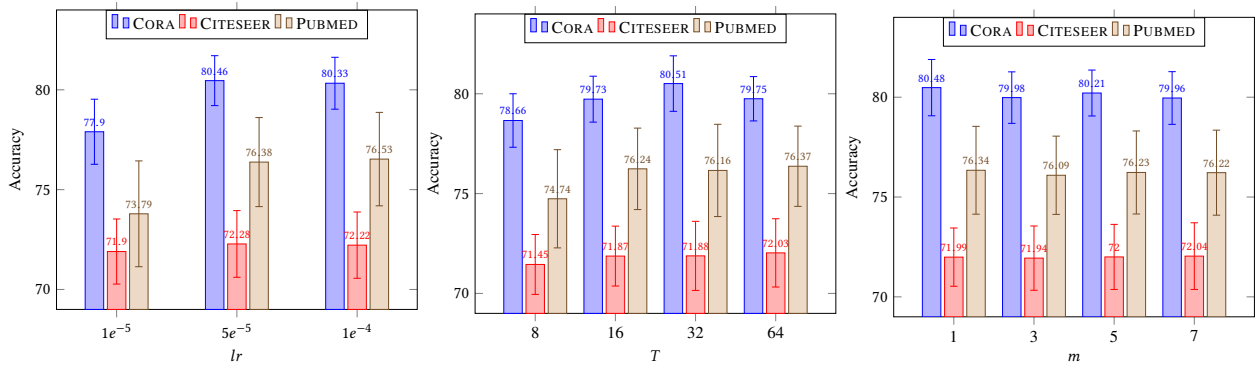

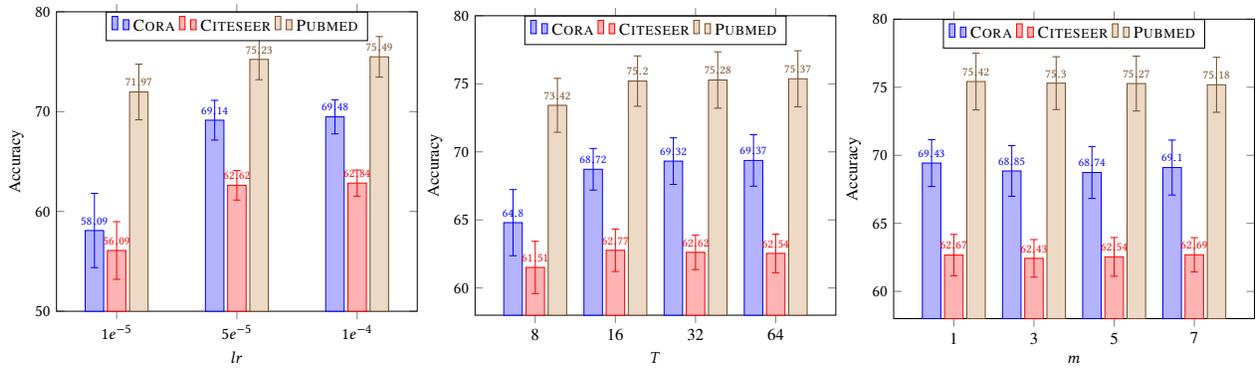
\begin{figure*}[!ht]
\centering
\centering
\resizebox{16.75cm}{!}{

\begin{tikzpicture}
\begin{axis}[
    ybar,
    enlarge x limits=0.25,
    legend style={at={(0.5,1)},
                anchor=north,legend columns=3},
    ylabel={Accuracy},
    xlabel={$lr$},
    symbolic x coords={$1e^{-5}$, $5e^{-5}$, $1e^{-4}$},
    xtick=data,
    ymin=50,ymax=80,
    nodes near coords={\scriptsize\pgfmathprintnumber\pgfplotspointmeta},
    ]
\addplot+[error bars/.cd, y dir=both,y explicit] coordinates {
($1e^{-5}$,58.09) +- (0.0, 3.72)
($5e^{-5}$,69.14) +- (0.0, 1.99) 
($1e^{-4}$,69.48) +- (0.0, 1.71)
};
\addplot+[error bars/.cd, y dir=both,y explicit] coordinates {
($1e^{-5}$,56.09) +- (0.0, 2.89)
($5e^{-5}$,62.62) +- (0.0, 1.48) 
($1e^{-4}$,62.84) +- (0.0, 1.31)
};
\addplot+[error bars/.cd, y dir=both,y explicit] coordinates {
($1e^{-5}$,71.97) +- (0.0, 2.79)
($5e^{-5}$,75.23) +- (0.0, 2.03) 
($1e^{-4}$,75.49) +- (0.0, 2.03)
};
\legend{\textsc{Cora},\textsc{Citeseer},\textsc{Pubmed}}
\end{axis}
\end{tikzpicture}

\begin{tikzpicture}
\begin{axis}[
    ybar,
    enlarge x limits=0.25,
    legend style={at={(0.5,1)},
                anchor=north,legend columns=3},
    ylabel={Accuracy},
    xlabel={$T$},
    symbolic x coords={8, 16, 32, 64},
    xtick=data,
    ymin=58,ymax=80,
    nodes near coords={\scriptsize\pgfmathprintnumber\pgfplotspointmeta},
    ]
\addplot+[error bars/.cd, y dir=both,y explicit] coordinates {
(8,64.80) +- (0.0, 2.43)
(16,68.72) +- (0.0, 1.53) 
(32,69.32) +- (0.0, 1.71)
(64,69.37)+- (0.0, 1.89)
};
\addplot+[error bars/.cd, y dir=both,y explicit] coordinates {
(8,61.51) +- (0.0, 1.92)
(16,62.77) +- (0.0, 1.56) 
(32,62.62) +- (0.0, 1.27)
(64,62.54)+- (0.0, 1.42)
};
\addplot+[error bars/.cd, y dir=both,y explicit] coordinates {
(8,73.42) +- (0.0, 1.98)
(16,75.20) +- (0.0, 1.85) 
(32,75.28) +- (0.0, 2.07)
(64,75.37)+- (0.0, 2.06)
};
\legend{\textsc{Cora},\textsc{Citeseer},\textsc{Pubmed}}
\end{axis}
\end{tikzpicture}

\begin{tikzpicture}
\begin{axis}[
    ybar,
    enlarge x limits=0.25,
    legend style={at={(0.5,1)},
                anchor=north,legend columns=3},
    ylabel={Accuracy},
    xlabel={$m$},
    symbolic x coords={1, 3, 5, 7},
    xtick=data,
    ymin=58,ymax=80,
    nodes near coords={\scriptsize\pgfmathprintnumber\pgfplotspointmeta},
    ]
\addplot+[error bars/.cd, y dir=both,y explicit] coordinates {
(1,69.43) +- (0.0, 1.72)
(3,68.85) +- (0.0, 1.86) 
(5,68.74) +- (0.0, 1.91)
(7,69.10)+- (0.0, 2.02)
};
\addplot+[error bars/.cd, y dir=both,y explicit] coordinates {
(1,62.67) +- (0.0, 1.52)
(3,62.43) +- (0.0, 1.38) 
(5,62.54) +- (0.0, 1.42)
(7,62.69)+- (0.0, 1.26)
};
\addplot+[error bars/.cd, y dir=both,y explicit] coordinates {
(1,75.42) +- (0.0, 2.09)
(3,75.30) +- (0.0, 1.94) 
(5,75.27) +- (0.0, 2.02)
(7,75.18)+- (0.0, 2.02)
};
\legend{\textsc{Cora},\textsc{Citeseer},\textsc{Pubmed}}
\end{axis}
\end{tikzpicture}

}
\caption{Effects of the Adam initial  learning rate $lr$ (left figure), the number $T$ of random walks sampled for each node (central figure), and the number $m$ of iterations in the routing process (right figure) on the validation sets in the inductive setting.}
\label{fig:EffectsRoutingIterations}
\end{figure*}

\subsection{Ablation analysis on the routing update}


\begin{table}[!ht]
\centering
\caption{Accuracy results on the \textsc{Cora} {validation} sets w.r.t each data split and each value $m > 1$ of routing iterations for the transductive and inductive settings. Regarding Algorithm \ref{alg:routing}  when $m > 1$, ``Ours'' denotes our update rule ($b_{i} \leftarrow \hat{\boldsymbol{\mathsf{u}}}_{\mathsf{v}_i}^{(i)}\cdot\boldsymbol{\mathsf{e}}_{\mathsf{v}}$), while ``Sab.'' denotes the update rule ($b_{i} \leftarrow b_{i} + \hat{\boldsymbol{\mathsf{u}}}_{\mathsf{v}_i}^{(i)}\cdot\boldsymbol{\mathsf{e}}_{\mathsf{v}}$) originally used by  \citet{sabour2017dynamic}.}
\resizebox{8.5cm}{!}{
\setlength{\tabcolsep}{0.25em}
\begin{tabular}{l|cc|cc|cc||cc|cc|cc}
\hline
\multirow{3}{*}{Split} & \multicolumn{6}{c||}{\textbf{Transductive}} & \multicolumn{6}{c}{\textbf{Inductive}}\\
\cline{2-13}
 & \multicolumn{2}{c|}{$m$=3} & \multicolumn{2}{c|}{$m$=5} & \multicolumn{2}{c||}{$m$=7} & \multicolumn{2}{c|}{$m$=3} & \multicolumn{2}{c|}{$m$=5} & \multicolumn{2}{c}{$m$=7}\\
\cline{2-13}
& Ours & Sab. & Ours & Sab. & Ours & Sab. & Ours & Sab. & Ours & Sab. & Ours & Sab.\\
\hline
1st & 80.1 & 80.1 & {80.2} & 79.6 & 79.7 & 79.3 & 70.2 & 70.3 & 70.2 & 69.2 & {70.6} & 68.3\\
2nd & 79.4 & 79.6 & {79.7} & 78.9 & {79.7} & 78.6 & {66.0} & 65.9 & 65.7 & 64.4 & 65.6 & 64.3 \\
3rd & 78.5 & 78.5 & {78.6} & {78.6} & 78.5 & 78.4 & 68.2 & 67.6 & 68.3 & 68.4 & {69.2} & 67.6 \\
4th & {81.3} & 80.8 & 81.1 & 80.1 & 81.1 & 79.3 & {66.5} & 66.3 & {66.5} & 65.4 & 66.4 & 65.9 \\
5th & {81.9} & 81.6 & 81.7 & 81.5 & 81.7 & 80.9 & 69.4 & 68.7 & {69.9} & 68.5 & 69.5 & 68.1 \\
6th & 78.6 & {79.0} & 78.8 & 78.7 & 78.7 & 78.0 & 66.7 & 67.1 & 66.7 & 66.2 & {67.5} & 65.3 \\
7th & 80.1 & 80.2 & {80.5} & 80.0 & 79.9 & 79.4 & {70.4} & 70.1 & {70.4} & 69.9 & {70.4} & 68.8 \\
8th & 81.8 & 82.1 & 82.1 & 81.5 & {82.3} & 81.2 & 69.6 & 69.0 & 68.7 & 67.8 & {69.7} & 67.5 \\
9th & 79.3 & 79.4 & {79.7} & 78.1 & 78.6 & 77.8 & 71.2 & 70.8 & 71.5 & 71.7 & {72.2} & 70.1 \\
10th & 78.8 & 79.3 & {79.7} & 78.9 & 79.4 & 78.7 & {70.3} & 69.7 & 69.5 & 68.8 & 69.9 & 68.3\\
\hline
Overall & 79.98 & \textbf{80.06} & \textbf{80.21} & 79.59 & \textbf{79.96} & 79.16 & \textbf{68.85} & 68.55 & \textbf{68.74} & 68.03 & \textbf{69.10} & 67.42\\
\hline
\end{tabular}
}
\label{tab:effectUpdate}
\end{table}

The routing process presented in Algorithm \ref{alg:routing} can be considered as an attention mechanism to compute the coupling coefficient $c_{i}$ which is used to weight the output of the $i$-th  capsule in the first layer.
\citet{sabour2017dynamic} use ($b_{i} \leftarrow b_{i} + \hat{\boldsymbol{\mathsf{u}}}_{\mathsf{v}_i}^{(i)}\cdot\boldsymbol{\mathsf{e}}_{\mathsf{v}}$) for the image classification task, but this might not be well-suited for graph-structured data because of the high order variant among different nodes. 
Therefore, we propose to use the new update rule ($b_{i} \leftarrow \hat{\boldsymbol{\mathsf{u}}}_{\mathsf{v}_i}^{(i)}\cdot\boldsymbol{\mathsf{e}}_{\mathsf{v}}$) as this new rule generally helps obtain a higher performance for each setup.
Table \ref{tab:effectUpdate} shows a comparison between the accuracy results of these two update rules on the \textsc{Cora} validation sets w.r.t each data split and the number $m$ ($m > 1$) of routing iterations.

\subsection{Effects of hyper-parameters}

Figures \ref{fig:EffectsTransductive}  and \ref{fig:EffectsRoutingIterations} presents effects of the Adam initial learning rate $lr$, the number $T$ of random walks sampled for each node and the number $m$ of iterations in the routing process on the validation sets in the transductive and inductive settings respectively. In these experiments, for the 10 data splits of each dataset, we apply the same value of one hyper-parameter and then tune other hyper-parameters.

We find that in general  using  $lr=1e^{-4}$ produces the top scores on the validation sets to both transductive and inductive settings. We also find that we generally obtain high accuracies with a high value of $T$ at either 32 or 64. However, there is an exception in the inductive setting,  where using $T=16$ produces the highest accuracy  on \textsc{Citeseer}. A possible reason might come from the fact that \textsc{Citeseer} is more sparse than \textsc{Cora} and \textsc{Pubmed}: the average number of neighbors per node on \textsc{Citeseer} is 1.4 which is substantially smaller than 2.0 on \textsc{Cora}  and 2.2 on  \textsc{Pubmed}.

Furthermore, using $m=1$ usually obtains the top performances in both the settings.
But we also note that the best configurations of hyper-parameters over 10 data splits are not always relied on using $m=1$.

\section{Conclusions and future work}
\label{sec:conclusion}

In this paper, we present a new unsupervised embedding model Caps2NE based on the capsule network to learn node embeddings from the graph-structured data.
Our proposed Caps2NE aims to effectively use context neighbors in random walks to infer plausible embeddings for target nodes.
Experimental results show that Caps2NE obtains state-of-the-art performances on benchmark datasets for the node classification task.

\section*{Acknowledgement}
This research was partially supported by the ARC Discovery Projects DP150100031 and DP160103934.

\bibliographystyle{ACM-Reference-Format}
\bibliography{references}


\begin{thebibliography}{26}


\ifx \showCODEN    \undefined \def \showCODEN     #1{\unskip}     \fi
\ifx \showDOI      \undefined \def \showDOI       #1{#1}\fi
\ifx \showISBNx    \undefined \def \showISBNx     #1{\unskip}     \fi
\ifx \showISBNxiii \undefined \def \showISBNxiii  #1{\unskip}     \fi
\ifx \showISSN     \undefined \def \showISSN      #1{\unskip}     \fi
\ifx \showLCCN     \undefined \def \showLCCN      #1{\unskip}     \fi
\ifx \shownote     \undefined \def \shownote      #1{#1}          \fi
\ifx \showarticletitle \undefined \def \showarticletitle #1{#1}   \fi
\ifx \showURL      \undefined \def \showURL       {\relax}        \fi
\providecommand\bibfield[2]{#2}
\providecommand\bibinfo[2]{#2}
\providecommand\natexlab[1]{#1}
\providecommand\showeprint[2][]{arXiv:#2}

\bibitem[\protect\citeauthoryear{Battaglia, Hamrick, Bapst, Sanchez-Gonzalez,
  Zambaldi, Malinowski, Tacchetti, Raposo, Santoro, Faulkner,
  et~al\mbox{.}}{Battaglia et~al\mbox{.}}{2018}]%
        {battaglia2018relational}
\bibfield{author}{\bibinfo{person}{Peter~W Battaglia},
  \bibinfo{person}{Jessica~B Hamrick}, \bibinfo{person}{Victor Bapst},
  \bibinfo{person}{Alvaro Sanchez-Gonzalez}, \bibinfo{person}{Vinicius
  Zambaldi}, \bibinfo{person}{Mateusz Malinowski}, \bibinfo{person}{Andrea
  Tacchetti}, \bibinfo{person}{David Raposo}, \bibinfo{person}{Adam Santoro},
  \bibinfo{person}{Ryan Faulkner}, {et~al\mbox{.}}}
  \bibinfo{year}{2018}\natexlab{}.
\newblock \showarticletitle{Relational inductive biases, deep learning, and
  graph networks}.
\newblock \bibinfo{journal}{\emph{arXiv preprint arXiv:1806.01261}}
  (\bibinfo{year}{2018}).
\newblock


\bibitem[\protect\citeauthoryear{Breitkreutz, Stark, Reguly, Boucher,
  Breitkreutz, Livstone, Oughtred, Lackner, Bähler, Wood, Dolinski, and
  Tyers}{Breitkreutz et~al\mbox{.}}{2008}]%
        {BioGRID:2008}
\bibfield{author}{\bibinfo{person}{Bobby-Joe Breitkreutz},
  \bibinfo{person}{Chris Stark}, \bibinfo{person}{Teresa Reguly},
  \bibinfo{person}{Lorrie Boucher}, \bibinfo{person}{Ashton Breitkreutz},
  \bibinfo{person}{Michael Livstone}, \bibinfo{person}{Rose Oughtred},
  \bibinfo{person}{Daniel Lackner}, \bibinfo{person}{Jürg Bähler},
  \bibinfo{person}{Valerie Wood}, \bibinfo{person}{Kara Dolinski}, {and}
  \bibinfo{person}{Mike Tyers}.} \bibinfo{year}{2008}\natexlab{}.
\newblock \showarticletitle{{The BioGRID interaction database: 2008 update}}.
\newblock \bibinfo{journal}{\emph{Nucleic acids research}}
  \bibinfo{volume}{36} (\bibinfo{year}{2008}), \bibinfo{pages}{D637--40}.
\newblock


\bibitem[\protect\citeauthoryear{Cai, Zheng, and Chang}{Cai
  et~al\mbox{.}}{2018}]%
        {cai2018comprehensive}
\bibfield{author}{\bibinfo{person}{Hongyun Cai}, \bibinfo{person}{Vincent~W
  Zheng}, {and} \bibinfo{person}{Kevin Chang}.}
  \bibinfo{year}{2018}\natexlab{}.
\newblock \showarticletitle{A comprehensive survey of graph embedding:
  problems, techniques and applications}.
\newblock \bibinfo{journal}{\emph{IEEE Transactions on Knowledge and Data
  Engineering}}  \bibinfo{volume}{30} (\bibinfo{year}{2018}),
  \bibinfo{pages}{1616--1637}.
\newblock


\bibitem[\protect\citeauthoryear{Chen, Perozzi, Al-Rfou, and Skiena}{Chen
  et~al\mbox{.}}{2018}]%
        {Chen180802590}
\bibfield{author}{\bibinfo{person}{Haochen Chen}, \bibinfo{person}{Bryan
  Perozzi}, \bibinfo{person}{Rami Al-Rfou}, {and} \bibinfo{person}{Steven
  Skiena}.} \bibinfo{year}{2018}\natexlab{}.
\newblock \showarticletitle{A Tutorial on Network Embeddings}.
\newblock \bibinfo{journal}{\emph{arXiv preprint arXiv:1808.02590}}
  (\bibinfo{year}{2018}).
\newblock


\bibitem[\protect\citeauthoryear{Cui, Henrickson, Ke, and Wang}{Cui
  et~al\mbox{.}}{2018}]%
        {Cui1802.07007}
\bibfield{author}{\bibinfo{person}{Zhiyong Cui}, \bibinfo{person}{Kristian
  Henrickson}, \bibinfo{person}{Ruimin Ke}, {and} \bibinfo{person}{Yinhai
  Wang}.} \bibinfo{year}{2018}\natexlab{}.
\newblock \showarticletitle{High-Order Graph Convolutional Recurrent Neural
  Network: A Deep Learning Framework for Network-Scale Traffic Learning and
  Forecasting}.
\newblock \bibinfo{journal}{\emph{arXiv preprint arXiv:1802.07007}}
  (\bibinfo{year}{2018}).
\newblock


\bibitem[\protect\citeauthoryear{Duran and Niepert}{Duran and Niepert}{2017}]%
        {duran2017learning}
\bibfield{author}{\bibinfo{person}{Alberto~Garcia Duran} {and}
  \bibinfo{person}{Mathias Niepert}.} \bibinfo{year}{2017}\natexlab{}.
\newblock \showarticletitle{{Learning Graph Representations with Embedding
  Propagation}}. In \bibinfo{booktitle}{\emph{NIPS}}.
  \bibinfo{pages}{5119--5130}.
\newblock


\bibitem[\protect\citeauthoryear{Fan, Chang, Hsieh, Wang, and Lin}{Fan
  et~al\mbox{.}}{2008}]%
        {Fan:2008}
\bibfield{author}{\bibinfo{person}{Rong-En Fan}, \bibinfo{person}{Kai-Wei
  Chang}, \bibinfo{person}{Cho-Jui Hsieh}, \bibinfo{person}{Xiang-Rui Wang},
  {and} \bibinfo{person}{Chih-Jen Lin}.} \bibinfo{year}{2008}\natexlab{}.
\newblock \showarticletitle{{LIBLINEAR: A Library for Large Linear
  Classification}}.
\newblock \bibinfo{journal}{\emph{Journal of Machine Learning Research}}
  \bibinfo{volume}{9} (\bibinfo{year}{2008}), \bibinfo{pages}{1871--1874}.
\newblock


\bibitem[\protect\citeauthoryear{Grover and Leskovec}{Grover and
  Leskovec}{2016}]%
        {Grover:2016}
\bibfield{author}{\bibinfo{person}{Aditya Grover} {and} \bibinfo{person}{Jure
  Leskovec}.} \bibinfo{year}{2016}\natexlab{}.
\newblock \showarticletitle{{Node2Vec: Scalable Feature Learning for
  Networks}}. In \bibinfo{booktitle}{\emph{SIGKDD}}. \bibinfo{pages}{855--864}.
\newblock


\bibitem[\protect\citeauthoryear{Guo, Xu, and Chen}{Guo et~al\mbox{.}}{2018}]%
        {guo2018spine}
\bibfield{author}{\bibinfo{person}{Junliang Guo}, \bibinfo{person}{Linli Xu},
  {and} \bibinfo{person}{Enhong Chen}.} \bibinfo{year}{2018}\natexlab{}.
\newblock \showarticletitle{SPINE: Structural Identity Preserved Inductive
  Network Embedding}.
\newblock \bibinfo{journal}{\emph{arXiv preprint arXiv:1802.03984}}
  (\bibinfo{year}{2018}).
\newblock


\bibitem[\protect\citeauthoryear{Hamilton, Ying, and Leskovec}{Hamilton
  et~al\mbox{.}}{2017}]%
        {hamilton2017inductive}
\bibfield{author}{\bibinfo{person}{William~L. Hamilton}, \bibinfo{person}{Rex
  Ying}, {and} \bibinfo{person}{Jure Leskovec}.}
  \bibinfo{year}{2017}\natexlab{}.
\newblock \showarticletitle{Inductive representation learning on large graphs}.
  In \bibinfo{booktitle}{\emph{NIPS}}. \bibinfo{pages}{1024--1034}.
\newblock


\bibitem[\protect\citeauthoryear{Jean, Cho, Memisevic, and Bengio}{Jean
  et~al\mbox{.}}{2015}]%
        {Jean2015}
\bibfield{author}{\bibinfo{person}{S{\'e}bastien Jean},
  \bibinfo{person}{Kyunghyun Cho}, \bibinfo{person}{Roland Memisevic}, {and}
  \bibinfo{person}{Yoshua Bengio}.} \bibinfo{year}{2015}\natexlab{}.
\newblock \showarticletitle{On Using Very Large Target Vocabulary for Neural
  Machine Translation}. In \bibinfo{booktitle}{\emph{ACL}}.
  \bibinfo{pages}{1--10}.
\newblock


\bibitem[\protect\citeauthoryear{Kingma and Ba}{Kingma and Ba}{2014}]%
        {kingma2014adam}
\bibfield{author}{\bibinfo{person}{Diederik Kingma} {and}
  \bibinfo{person}{Jimmy Ba}.} \bibinfo{year}{2014}\natexlab{}.
\newblock \showarticletitle{Adam: A method for stochastic optimization}.
\newblock \bibinfo{journal}{\emph{arXiv preprint arXiv:1412.6980}}
  (\bibinfo{year}{2014}).
\newblock


\bibitem[\protect\citeauthoryear{Kipf and Welling}{Kipf and Welling}{2017}]%
        {kipf2017semi}
\bibfield{author}{\bibinfo{person}{Thomas~N. Kipf} {and} \bibinfo{person}{Max
  Welling}.} \bibinfo{year}{2017}\natexlab{}.
\newblock \showarticletitle{{Semi-Supervised Classification with Graph
  Convolutional Networks}}. In \bibinfo{booktitle}{\emph{ICLR}}.
\newblock


\bibitem[\protect\citeauthoryear{Mahoney}{Mahoney}{2011}]%
        {mahoney2011large}
\bibfield{author}{\bibinfo{person}{Matt Mahoney}.}
  \bibinfo{year}{2011}\natexlab{}.
\newblock \bibinfo{title}{Large text compression benchmark}.
\newblock \bibinfo{howpublished}{http://www.mattmahoney.net/text/text.html}.
\newblock


\bibitem[\protect\citeauthoryear{Mikolov, Sutskever, Chen, Corrado, and
  Dean}{Mikolov et~al\mbox{.}}{2013}]%
        {MikolovSCCD13nips}
\bibfield{author}{\bibinfo{person}{Tomas Mikolov}, \bibinfo{person}{Ilya
  Sutskever}, \bibinfo{person}{Kai Chen}, \bibinfo{person}{Gregory~S. Corrado},
  {and} \bibinfo{person}{Jeffrey Dean}.} \bibinfo{year}{2013}\natexlab{}.
\newblock \showarticletitle{{Distributed Representations of Words and Phrases
  and their Compositionality}}. In \bibinfo{booktitle}{\emph{NIPS}}.
  \bibinfo{pages}{3111--3119}.
\newblock


\bibitem[\protect\citeauthoryear{Namata, London, Getoor, and Huang}{Namata
  et~al\mbox{.}}{2012}]%
        {namata:mlg12}
\bibfield{author}{\bibinfo{person}{Galileo~Mark Namata}, \bibinfo{person}{Ben
  London}, \bibinfo{person}{Lise Getoor}, {and} \bibinfo{person}{Bert Huang}.}
  \bibinfo{year}{2012}\natexlab{}.
\newblock \showarticletitle{{Query-driven Active Surveying for Collective
  Classification}}. In \bibinfo{booktitle}{\emph{Workshop on Mining and
  Learning with Graphs}}.
\newblock


\bibitem[\protect\citeauthoryear{Nguyen, Nguyen, and Phung}{Nguyen
  et~al\mbox{.}}{2020}]%
        {Nguyen2019SANNE}
\bibfield{author}{\bibinfo{person}{Dai~Quoc Nguyen}, \bibinfo{person}{Tu~Dinh
  Nguyen}, {and} \bibinfo{person}{Dinh Phung}.}
  \bibinfo{year}{2020}\natexlab{}.
\newblock \showarticletitle{{A Self-Attention Network based Node Embedding
  Model}}. In \bibinfo{booktitle}{\emph{ECML-PKDD}}.
\newblock


\bibitem[\protect\citeauthoryear{Perozzi, Al-Rfou, and Skiena}{Perozzi
  et~al\mbox{.}}{2014}]%
        {Perozzi:2014}
\bibfield{author}{\bibinfo{person}{Bryan Perozzi}, \bibinfo{person}{Rami
  Al-Rfou}, {and} \bibinfo{person}{Steven Skiena}.}
  \bibinfo{year}{2014}\natexlab{}.
\newblock \showarticletitle{{DeepWalk: Online Learning of Social
  Representations}}. In \bibinfo{booktitle}{\emph{SIGKDD}}.
  \bibinfo{pages}{701--710}.
\newblock


\bibitem[\protect\citeauthoryear{Sabour, Frosst, and Hinton}{Sabour
  et~al\mbox{.}}{2017}]%
        {sabour2017dynamic}
\bibfield{author}{\bibinfo{person}{Sara Sabour}, \bibinfo{person}{Nicholas
  Frosst}, {and} \bibinfo{person}{Geoffrey~E Hinton}.}
  \bibinfo{year}{2017}\natexlab{}.
\newblock \showarticletitle{Dynamic routing between capsules}. In
  \bibinfo{booktitle}{\emph{NIPS}}. \bibinfo{pages}{3859--3869}.
\newblock


\bibitem[\protect\citeauthoryear{Sen, Namata, Bilgic, Getoor, Galligher, and
  Eliassi-Rad}{Sen et~al\mbox{.}}{2008}]%
        {sen2008collective}
\bibfield{author}{\bibinfo{person}{Prithviraj Sen}, \bibinfo{person}{Galileo
  Namata}, \bibinfo{person}{Mustafa Bilgic}, \bibinfo{person}{Lise Getoor},
  \bibinfo{person}{Brian Galligher}, {and} \bibinfo{person}{Tina Eliassi-Rad}.}
  \bibinfo{year}{2008}\natexlab{}.
\newblock \showarticletitle{Collective classification in network data}.
\newblock \bibinfo{journal}{\emph{AI magazine}} \bibinfo{volume}{29},
  \bibinfo{number}{3} (\bibinfo{year}{2008}), \bibinfo{pages}{93}.
\newblock


\bibitem[\protect\citeauthoryear{Veli{\v{c}}kovi{\'{c}}, Cucurull, Casanova,
  Romero, Li{\`{o}}, and Bengio}{Veli{\v{c}}kovi{\'{c}} et~al\mbox{.}}{2018}]%
        {velickovic2018graph}
\bibfield{author}{\bibinfo{person}{Petar Veli{\v{c}}kovi{\'{c}}},
  \bibinfo{person}{Guillem Cucurull}, \bibinfo{person}{Arantxa Casanova},
  \bibinfo{person}{Adriana Romero}, \bibinfo{person}{Pietro Li{\`{o}}}, {and}
  \bibinfo{person}{Yoshua Bengio}.} \bibinfo{year}{2018}\natexlab{}.
\newblock \showarticletitle{{Graph Attention Networks}}. In
  \bibinfo{booktitle}{\emph{ICLR}}.
\newblock


\bibitem[\protect\citeauthoryear{Wang, Huang, Zhao, Zhang, Zhao, and Lee}{Wang
  et~al\mbox{.}}{2018}]%
        {Wang:2018:BCE}
\bibfield{author}{\bibinfo{person}{Jizhe Wang}, \bibinfo{person}{Pipei Huang},
  \bibinfo{person}{Huan Zhao}, \bibinfo{person}{Zhibo Zhang},
  \bibinfo{person}{Binqiang Zhao}, {and} \bibinfo{person}{Dik~Lun Lee}.}
  \bibinfo{year}{2018}\natexlab{}.
\newblock \showarticletitle{{Billion-scale Commodity Embedding for E-commerce
  Recommendation in Alibaba}}. In \bibinfo{booktitle}{\emph{SIGKDD}}.
  \bibinfo{pages}{839--848}.
\newblock


\bibitem[\protect\citeauthoryear{Yang, Cohen, and Salakhutdinov}{Yang
  et~al\mbox{.}}{2016}]%
        {Yang:2016planetoid}
\bibfield{author}{\bibinfo{person}{Zhilin Yang}, \bibinfo{person}{William~W.
  Cohen}, {and} \bibinfo{person}{Ruslan Salakhutdinov}.}
  \bibinfo{year}{2016}\natexlab{}.
\newblock \showarticletitle{{Revisiting Semi-supervised Learning with Graph
  Embeddings}}. In \bibinfo{booktitle}{\emph{ICML}}. \bibinfo{pages}{40--48}.
\newblock


\bibitem[\protect\citeauthoryear{Ying, He, Chen, Eksombatchai, Hamilton, and
  Leskovec}{Ying et~al\mbox{.}}{2018}]%
        {Ying:2018:GCN}
\bibfield{author}{\bibinfo{person}{Rex Ying}, \bibinfo{person}{Ruining He},
  \bibinfo{person}{Kaifeng Chen}, \bibinfo{person}{Pong Eksombatchai},
  \bibinfo{person}{William~L. Hamilton}, {and} \bibinfo{person}{Jure
  Leskovec}.} \bibinfo{year}{2018}\natexlab{}.
\newblock \showarticletitle{{Graph Convolutional Neural Networks for Web-Scale
  Recommender Systems}}. In \bibinfo{booktitle}{\emph{SIGKDD}}.
  \bibinfo{pages}{974--983}.
\newblock


\bibitem[\protect\citeauthoryear{Zafarani and Liu}{Zafarani and Liu}{2009}]%
        {ZafaraniLiu:2009}
\bibfield{author}{\bibinfo{person}{R. Zafarani} {and} \bibinfo{person}{H.
  Liu}.} \bibinfo{year}{2009}\natexlab{}.
\newblock \bibinfo{title}{Social Computing Data Repository at {ASU}}.
\newblock \bibinfo{howpublished}{http://socialcomputing.asu.edu}.
\newblock


\bibitem[\protect\citeauthoryear{Zhang, Yin, Zhu, and Zhang}{Zhang
  et~al\mbox{.}}{2020}]%
        {zhang2020network}
\bibfield{author}{\bibinfo{person}{Daokun Zhang}, \bibinfo{person}{Jie Yin},
  \bibinfo{person}{Xingquan Zhu}, {and} \bibinfo{person}{Chengqi Zhang}.}
  \bibinfo{year}{2020}\natexlab{}.
\newblock \showarticletitle{Network representation learning: A survey}.
\newblock \bibinfo{journal}{\emph{IEEE Transactions on Big Data}}
  (\bibinfo{year}{2020}), \bibinfo{pages}{3--28}.
\newblock


\end{thebibliography}

\end{document}